# A Text to Speech (TTS) System with English to Punjabi Conversion

Prabhsimran Singh
Department of Computer Science & Engineering
Guru Nanak Dev University, Amritsar
prabh_singh32@yahoo.com

Amritpal Singh
Department of Computer Science & Engineering
Guru Nanak Dev University, Amritsar
amritsinghmand@gmail.com

*Abstract*—The paper aims to show how an application can be developed that converts the English language into the Punjabi Language, and the same application can convert the Text to Speech(TTS) i.e. pronounce the text. This application can be really beneficial for those with special needs.

## I. INTRODUCTION

Now days the main aim of technology is to help people in their day to day requirements and facilitate their work. People of all age categories from young to old use computers, even those with visual disabilities.

All this has been made possible with the development of systems which can recognize text and speech, this makes blind and visually impaired people to use computer systems without the help of any other person. But if we combine these systems with language translation systems it gives an added advantage to those who are not so familiar with English language[1].

## II. TEXT TO SPEECH

A Text to Speech (TTS) system is a computer system that converts text into speech, i.e. read the text automatically when asked to do so[2]. This system is a combination of both hardware and software. Generally a Text (Sentence) is composed of collection of words, while words are combination of alphabets arranged in a meaningful way. The work on (TTS) systems goes back to 1779, when the Danish scientist Christian Kratzenstein, build a human vocal model that can produce five long vowel sounds i.e. [a], [e], [i], [o] and [u].

Generally a TTS system consists of 5 fundamental phases:

    a. Text Analysis and Detection
    b. Text Normalization and Linearization
    c. Phonetic Analysis
    d. Prosodic Modeling and Intonation
    e. Acoustic Processing

The Text to be converted is passed through these phases step by step in order to obtain the Speech. Figure 1, shows diagrammatical representation of these phases.

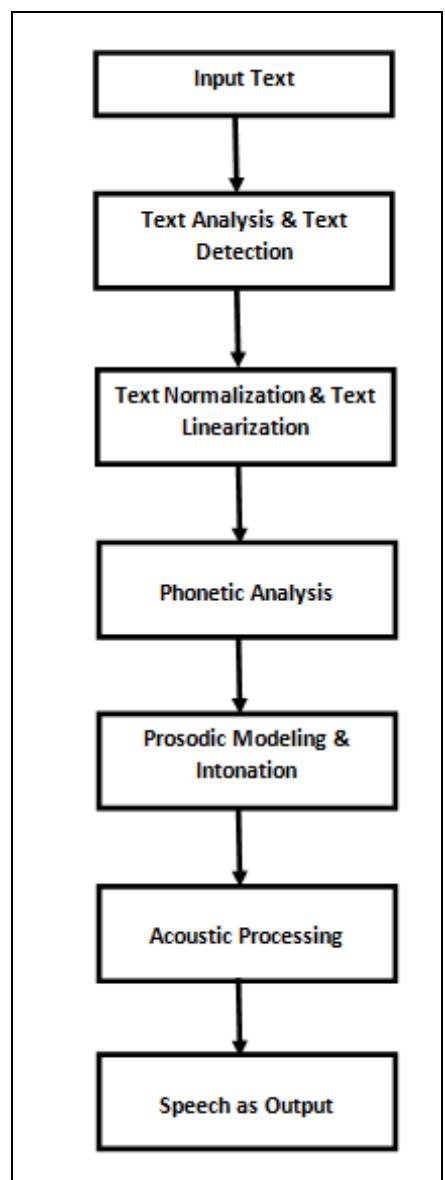

Fig. 1 "Overview of TTS"[2]

## III. LANGUAGE TRANSLATOR

A language translator is a system that converts the input text from one language to another. There are various translators available on the internet, with most famous being the Google Translate[3]. It support multiple language ranging over different regionals as well as national languages. Figure 2, shows a screenshot of Google Translate, where English is being converted to Spanish language. The aim of this paper is to convert English Language to Punjabi Language, and to give additional functionality of pronouncing the text i.e. Text to speech (TTS)..

## IV. LITURATURE REVIEW AND MOTIVATION

All around the world various systems has been developed which translate one language to another and that covert text to speech. Some of them which act as motivation for our system are as following:

**A.** English language to Arabic language convertor by Ameera Al-Rehili, Dalal Al-Juhani, Maha Al-Maimani and Munir Ahmed. It converts English Language to Arabic Language and Vice Versa. Moreover has the ability to pronounce English as well as Arabic Text [1].

**B.** A system that converts Hindi language to Telugu Language and vice versa proposed and implemented by Lakshmi Sahu. This system has the ability to pronounce hindi and telugu text in male voice and female voice [4].

**C.** "Text to Speech: A Simple Tutorial" by D. Sasirekha and E. Chandra, shows what all is required to develop a system that has capabilities to convert Text to Speech. He paper discuses the various TTS systems and continuous improvement that is taking place in this field [2].

## V. PROPOSED SYSTEM

The aim of the proposed system is to develop a system that has capability to perform Translation (English Language to Punjabi Language), Converting (Text to Speech i.e. read the Text aloud). The proposed system will be developed in Microsoft Visual Web Developer Express edition 2010, with C#.Net as front end while SQL Server 2008 as back end [5]. The System thus developed can be used as a web application or desktop application by deploying it as local host. The system proposed her is developed for a small domain of English words.

### A. Requirements for the proposed system

The requirements can be classified as hardware and software. The Minimum Hardware requirements include Dual Core Processor, 2GB Ram, and 500 mb Hard disk. The minimum Software requirements are Microsoft Windows XP service pack 3 or higher, Microsoft .Net framework 4.0.

### B. Architecure of Proposed System

For each English words all possible Punjabi words substitutions are stored in the database, and infact this creates one of the major problem ambiguity i.e. for one English word there is more than one substitutions. Table 1, shows some words with multiple substitutions. So proper rules need to be applied to check which word fit's best in the sentence, while conversion of English to Punjabi.

Table 1, "Ambiguity in Punjabi for English words"

| English | Punjabi |
| --- | --- |
| You | ਤੁਸੀ |
| You | ਤੂੰ |
| Your | ਤੇਰਾ |
| Your | ਤੁਹਾਡਾ |

## VI. WORKING OF THE PROPOSED SYSTEM

The working of the entire system can be broken down into two main parts i.e. Translation and Text to Speech(TTS).

### A. Translation

The Figure 3, shows the User interface of the system. Whenever button to convert the text is pressed, the entire sentence/string is parsed(Divided) into single words/units, and then corresponding word/sound in the database is searched which satisfy certain set of rules for each and every words/units and is replaced. This process is repeated till the entire sentence is processed/converted. Figure 4, shows the entire process in step by step fashion.

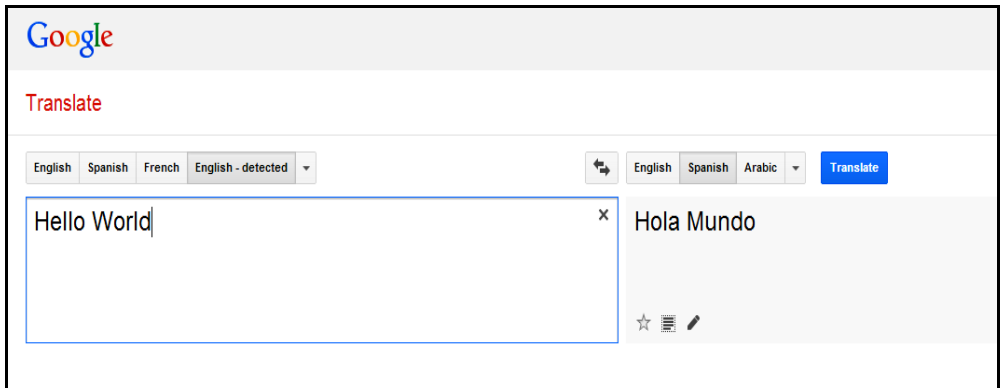

Fig 2. "Screenshot of Google Translate [3]"

## B. Detailed working of the translation System

Suppose we want to translate an English sentence **"Who did this?"** into Punjabi Language. First of all the entire string will be parsed, and will result in 4 independent unit/words. Then for each english word corresponding Punjabi word will be selected from the database as shown in Table 2.

Now if these words are combined to form a Punjabi sentence, it will result in *"ਕਿਸ ਨੇ ਕੀਤਾ ਇਹ?"*, which is grammatically incorrect. Here the rules come in play, and rearrangement of sentence take place to make it grammatically correct. Hence the final result will be *"ਇਹ ਕਿਸ ਨੇ ਕੀਤਾ?"*. The description is shown in figure 5.

Table 2, "Punjabi words for corresponding English words"

| English | Punjabi |
|---------|---------|
| Who | ਕਿਸ ਨੇ |
| did | ਕੀਤਾ |
| this | ਇਹ |
| ? | ? |

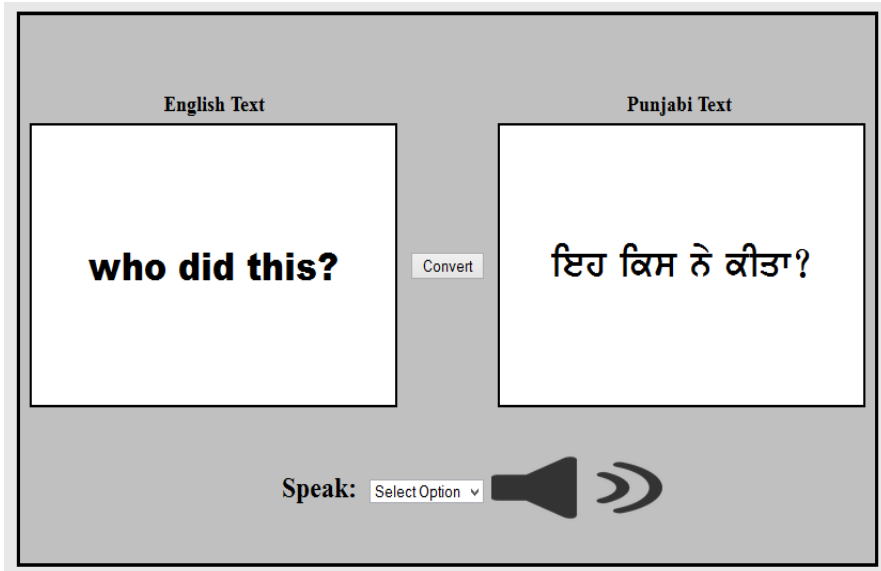

Fig 3, "User Interface of the System"

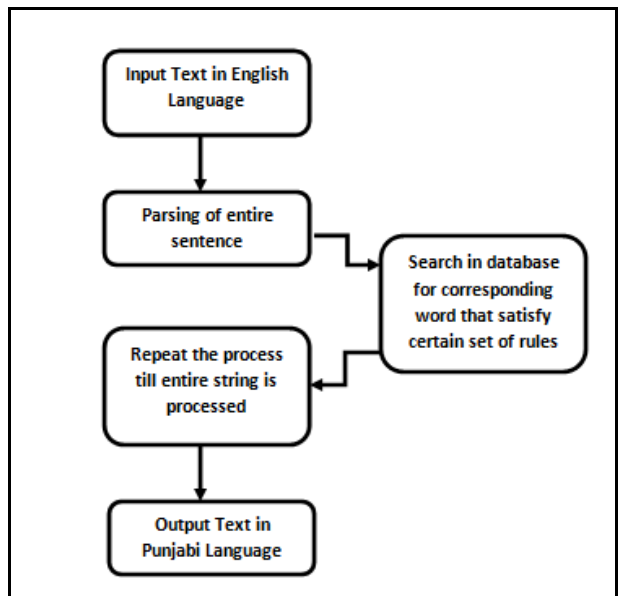

Fig 4, "Stepwise process of Language Translation"

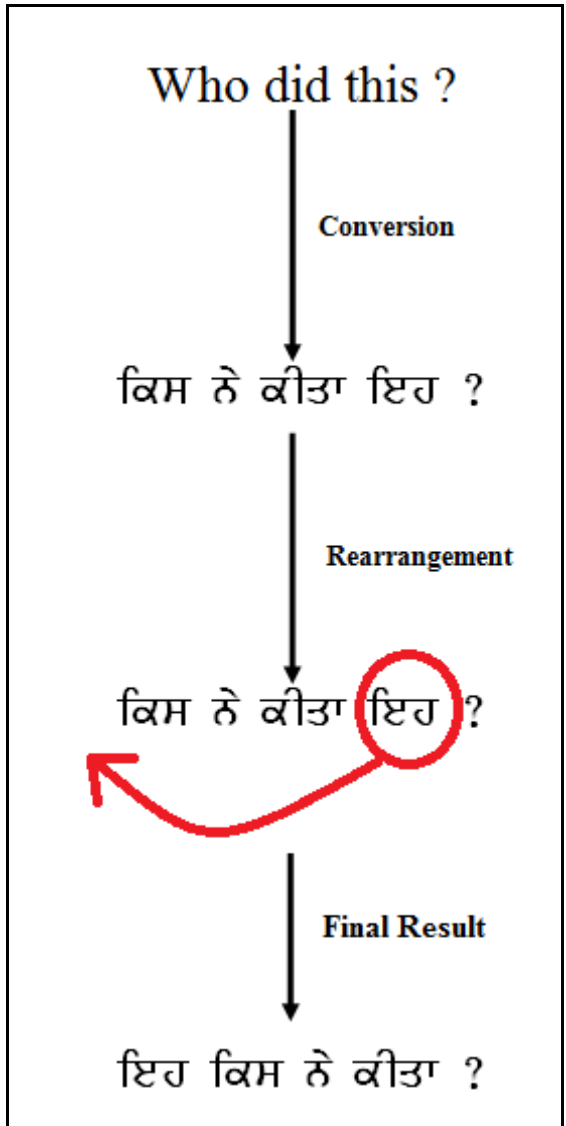

Fig 5, "Detail Working of Translation Process"

*C. Text to Speech(TTS)*

To convert Text to Speech we use the inbuilt .Net Library/Class "System.Speech.Synthesis"[6]. The Code for this is shown is figure 6. The System is coded to speak both English as well as Punjabi languages. When we select a language and press the Speak (picture) button the text is read aloud. The system accurately pronounces the Punjabi language. The working interface of TTS is shown in the figure 7 and 8.

```
//create voice object
SpVoice voice = new SpVoice();
//have it speak the text in out textbox
voice.Speak(txt1.Text);
```

Fig 6, "Code of TTS [6]"

**VII. COMPARISON OF PROPOSED SYSTEM WITH FEW OTHER SYSTEMS**

In this section the proposed system will be compared with few other systems based on various parameters like language used, database used etc as shown in table 3.

Table 3, "Comparison of Different System"

|  | Proposed System | System Proposed by Ameera Al-Rehili and group[1] | System Proposed by Lakhsmi Sahu[4] |
|---|---|---|---|
| **Front End (Language Used)** | C#.Net | Vb.Net | C#.Net |
| **Back End (Database)** | SQL Server 2008 | MS Access 2007 | --- |
| **Language Conversion Efficiency** | High | Moderate | High |
| **Text to Speech (TTS)** | Yes | Yes | Yes |
| **TTS other than English** | Yes (Punjabi) | Yes (Arabic) | Yes (Hindi & Telugu) |
| **Pronunciation Efficiency** | High | High | High |
| **Web / Desktop Application** | Both | Desktop | Desktop |

**VIII. CONCLUSION AND FUTURE WORK**

In this paper we presented a system that can convert the English sentence into corresponding Punjabi sentence, as well as read the text aloud. The system is capable to be used as a web application or desktop application (deploying it as local host), hence making it more flexible. This system can be really helpful for those who are not so familiar with English language as well as for those with visually disabilities.

In Future, various advancements can be made to the system like adding the facility of Speech to Text (STT) so that user do not have to enter text manually, it can be done simply by speaking only. Current system can only supports the conversion of English language to Punjabi language, but we can add support for conversion of more languages to it so that it can become more useful. Moreover adding the facility of inbuilt dictionary i.e. giving meaning of different words available in the sentence.

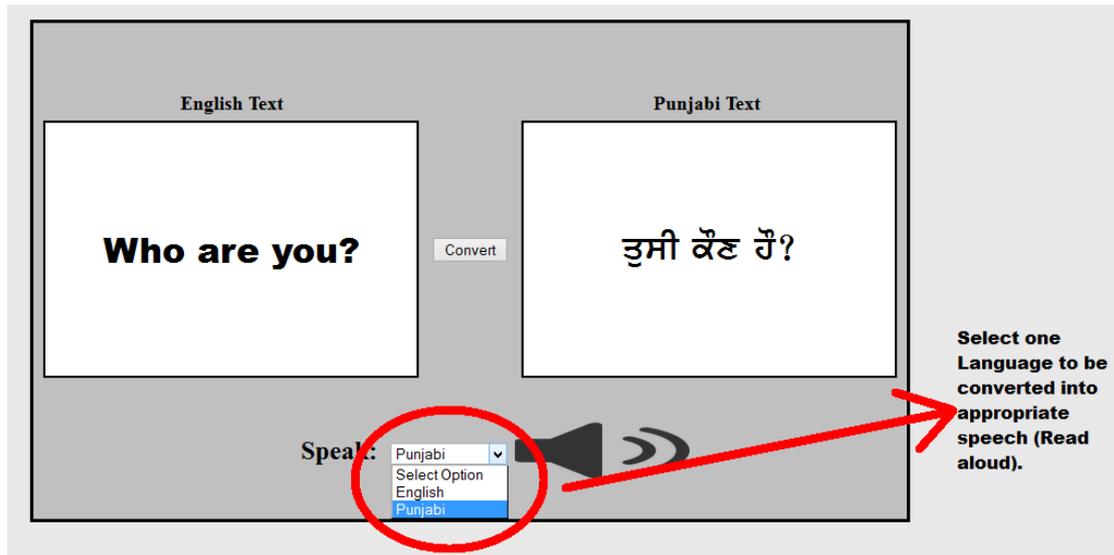

Fig 7, "Select Language Option"

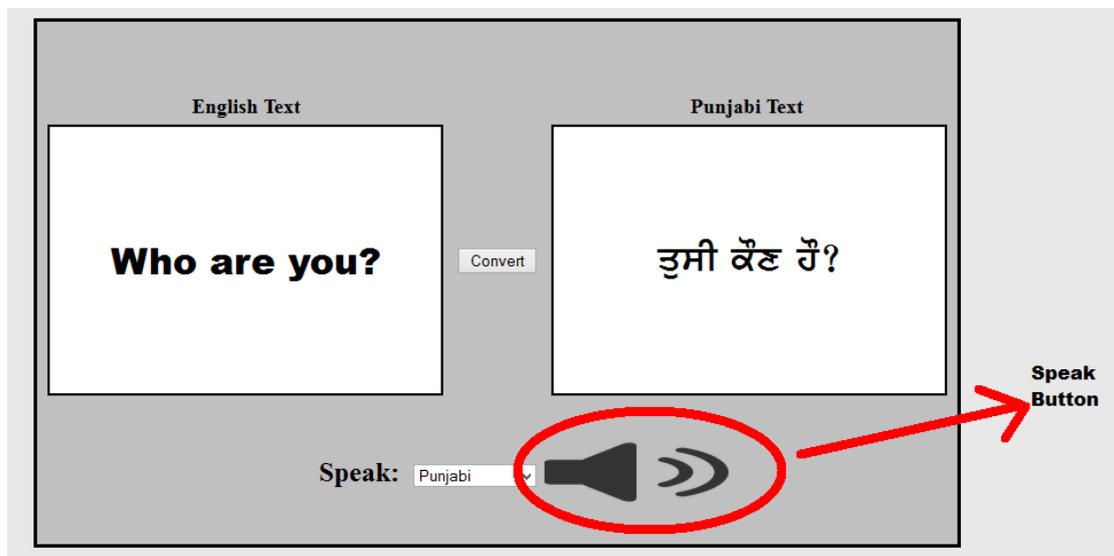

Fig 8, "Speak Button"